\documentclass[10pt,twocolumn,letterpaper]{article}

\usepackage{ijcb}
\usepackage{times}
\usepackage{epsfig}
\usepackage{graphicx}
\usepackage{amsmath}
\usepackage{amssymb}

\ijcbfinalcopy % *** Uncomment this line for the final submission

\begin{document}

%%%%%%%%% TITLE
\title{CLIP4Sketch: Enhancing Sketch to Mugshot Matching through Dataset Augmentation using Diffusion Models}

\author{Kushal Kumar Jain\\
IIIT-Hyderabad\\
Gachibowli, Hyderabad\\
{\tt\small kushal.kumar@reserarch.iiit.ac.in}
\and
Steven Grosz\\
Michigan State University\\
East Lansing, MI\\
{\tt\small groszste@msu.edu}
\and
Anoop M. Namboodiri\\
IIIT-Hyderabad\\
Gachibowli, Hyderabad\\
{\tt\small anoop@iiit.ac.in}
\and
Anil K. Jain\\
Michigan State University\\
East Lansing, MI\\
{\tt\small jain@msu.edu}
}

\maketitle
\thispagestyle{empty}

%%%%%%%%% ABSTRACT

\begin{abstract}

   Forensic sketch-to-mugshot matching is a challenging task in face recognition, primarily hindered by the scarcity of annotated forensic sketches and the modality gap between sketches and photographs. To address this, we propose CLIP4Sketch, a novel approach that leverages diffusion models to generate a large and diverse set of sketch images, which helps in enhancing the performance of face recognition systems in sketch-to-mugshot matching. Our method utilizes Denoising Diffusion Probabilistic Models (DDPMs) to generate sketches with explicit control over identity and style. We combine CLIP and Adaface embeddings of a reference mugshot, along with textual descriptions of style, as the conditions to the diffusion model. We demonstrate the efficacy of our approach by generating a comprehensive dataset of sketches corresponding to mugshots and training a face recognition model on our synthetic data. Our results show significant improvements in sketch-to-mugshot matching accuracy over training on an existing, limited amount of real face sketch data, validating the potential of diffusion models in enhancing the performance of face recognition systems across modalities. We also compare our dataset with datasets generated using GAN-based methods to show its superiority.  
\end{abstract}

%%%%%%%%% BODY TEXT
\section{Introduction}
Facial recognition technology has witnessed remarkable advancements in recent years \cite{Deng2018ArcFaceAA,Kim2022AdaFaceQA,Kim2023DCFaceSF}, yet challenges persist in accurately matching hand-drawn forensic sketches to corresponding mugshot photographs. This problem holds critical significance across various domains, including law enforcement, surveillance, and forensic investigations. Forensic sketch-to-mugshot matching presents inherent challenges stemming from the substantial discrepancy between these two data modalities. Forensic sketches, which are often incomplete artistic interpretations based on witness descriptions, exhibit stark differences from photographic mugshots captured under controlled conditions. Traditional approaches to this problem have primarily relied on discriminative frameworks that attempt to learn a shared subspace or extract specific features to enhance similarity between sketches and photos of the same identity while increasing dissimilarity across different identities \cite{Klare2011MatchingFS,klum}. However, the efficacy of such methods has been fundamentally limited by the need for more annotated forensic sketch data available for training and evaluation. The lack of sketch data has hampered the ability to assess method performance properly and limited the potential of local feature-based discriminant analysis techniques and recognition systems. Recently, the growing concerns regarding the legal and ethical implications of using authentic data in facial recognition (FR) training, coupled with the practical challenges in assembling large, diverse face datasets, have spurred research into the potential of synthetic data as a viable alternative to privacy-sensitive real world face data \cite{BOUTROS2023104688}.\\ 
To address the issue of scarcity of data, we propose a novel approach leveraging diffusion models for generating sketch images with large variations to enhance the performance of face recognition systems in sketch-to-mugshot matching. Diffusion models, characterized by their ability to generate high-quality samples from complex data distributions, offer a promising avenue for bridging the gap between sketches and photographs. Diffusion models \cite{Ho2020DenoisingDP,Song2020DenoisingDI,dhariwal2021diffusion} have made significant progress in text-guided image generation \cite{Ramesh2022HierarchicalTI,Saharia2022PhotorealisticTD, Rombach2021HighResolutionIS}, which has helped them garner much attention in the last few years. Before the advent of diffusion models, face sketch generation techniques predominantly employed Generative Adversarial Networks (GANs) \cite{Goodfellow2014GenerativeAN} trained on available datasets of photo-sketch pairs. However, GANs are susceptible to issues such as mode collapse, training instability, and limited generalization beyond the training data distribution, making them less suitable for the personalized generation of synthetic sketch datasets from mugshot images.\\
Our approach to match forensic sketches and mugshots uses traditional face recognition models trained on synthetic data generated using diffusion models. Inspired by IP-Adapter \cite{Ye2023IPAdapterTC} and other similar approaches \cite{li2023photomaker}, we guide our generation with explicit control over both identity and style by using mugshot image embeddings as prompts. We aggregated data from multiple publicly available sketch datasets \cite{Fan2021FS2K, 4624272, klum, tufts} and integrated ControlNet for spatial control and CLIP, along with Adaface, for identity control. We call this generation pipeline CLIP4Sketch. \\
In a field where accuracy is paramount, having control over the generated sketches is essential. CLIP4Sketch offers a solution to this issue, providing researchers with a powerful tool for improving face recognition technology. The CLIP4Sketch pipeline leverages the robust feature extraction capabilities of the CLIP encoder and the high-fidelity embeddings of Adaface to create a diverse set of sketch and mugshot image pairs. This synthetic dataset was then employed to finetune an AdaFace \cite{Kim2022AdaFaceQA} model, optimizing it to improve performance in sketch-to-face matching. Through rigorous training and testing, we validated the effectiveness of our dataset in enhancing the accuracy and reliability of the face recognition model. The results show the potential of our approach in significantly advancing the field of sketch-to-face recognition, demonstrating the practical benefits of the CLIP4Sketch pipeline in real-world applications. The contributions of this work can be summarised as :  \\
\begin{itemize}
    \item A controllable latent diffusion model, CLIP4Sketch, that uses text and image embeddings as conditions for highly realistic and diverse synthetic sketch generation.
    \item A synthetic dataset containing 27k unique identities with 4 hand-drawn style sketches and 4 software generated style sketches.
    \item Improvements in state-of-the-art face recognition systems toward mugshot to sketch matching when utilizing our synthetic data for training.  
\end{itemize}

\begin{figure*}[t]
% \begin{center}
   \includegraphics[width=1\linewidth]{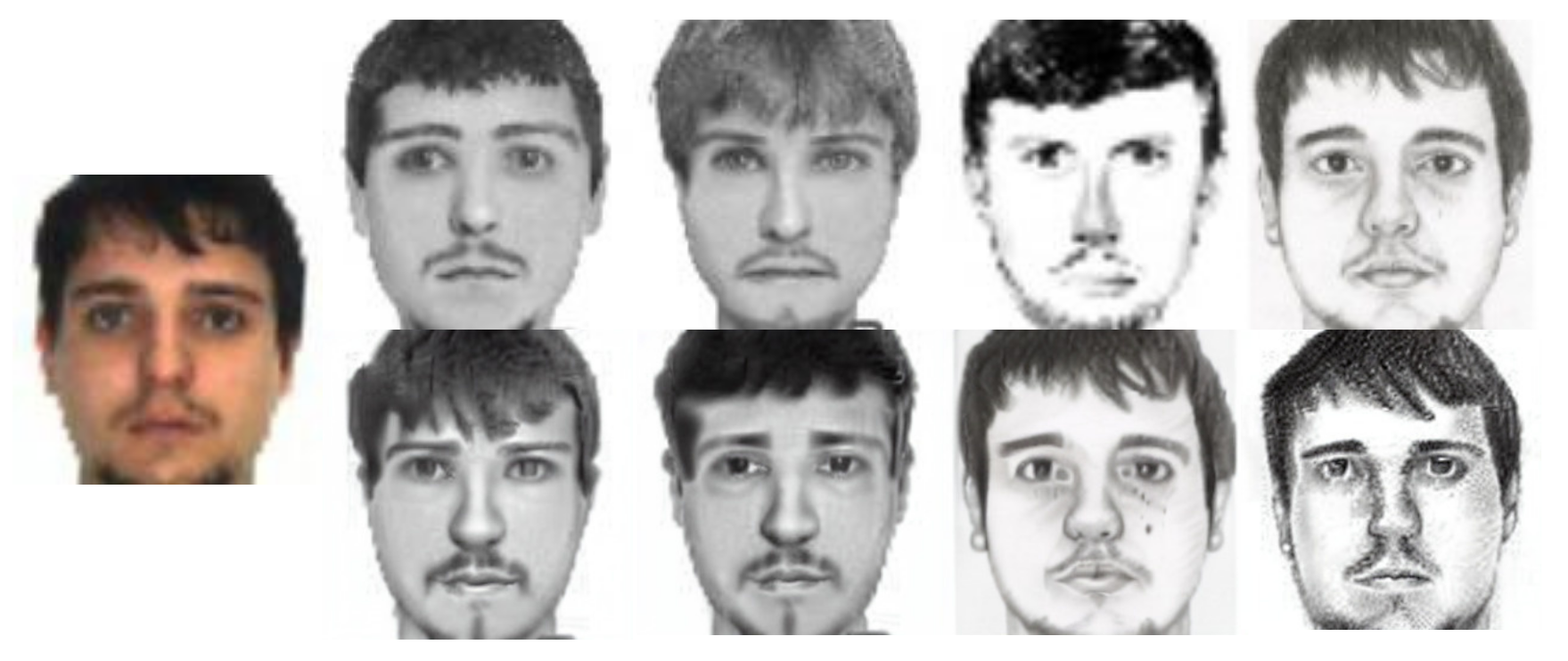}

% \end{center}
   \caption{Qualitative comparison of our CLIP4Sketch generated sketches and real sketches. The second row is generated by our method, while the first row contains examples from existing datasets PRIP-Composites \cite{klum} and CUHK\cite{4624272}. The text prompts used to generate the images in the second row using the proposed CLIP4Sketch model are, in order from left to right, “a viewed software-generated sketch of a face”, “a software-generated sketch of a face”, “a viewed hand-drawn sketch of a face”, and “a hand-drawn sketch of a face”. The leftmost image is the input identity. } 
\label{fig:qualitative}
\end{figure*}
\section{Related Works}
In this section we describe a few traditional methods for face to sketch matching and sketch generation.
Sketch recognition has been a subject of extensive research, but forensic sketches present larger modality gaps due to inherent errors made by witnesses and lack of sufficient data. Methods for photo to sketch matching can be broadly divided into two categories: discriminative or generative. Discriminative methods try to reduce the modality gap between photos and sketches by learning a shared subspace or extracting specific features that enhance the similarity within the same identity while increasing dissimilarity between different identities. Generative methods try to generate sketches from photos or vice-versa and then use traditional face-recognition algorithms to solve the problem. Most of these methods suffer from insufficient data and hence are unable to reliably perform well on other datasets.\\
\subsection{Photo to Sketch Matching}
Traditionally for photo to sketch matching hand-crafted local features have been used followed by dimensionality reduction and discriminant analysis. A few such methods are 
Partial Linear Squares (PLS) \cite{Sharma2011BypassingSP}, Coupled Information-Theoretic Projection (CITP) \cite{Zhang2011CoupledIE}, Local Feature-based Discriminant Analysis (LFDA) \cite{Klare2011MatchingFS} and Canonical Correlation Analysis (CCA) \cite{Ouyang2014CrossModalFM}. Additionally, component-based representation techniques \cite{Han2013MatchingCS} or Self-Similarity Descriptor (SSD) dictionary \cite{Mittal2014RecognizingCS}, Multi-scale Circular Weber's Local Descriptor (MCWLD) and other similar approaches \cite{Bhatt2012MemeticallyOM, Peng2015GraphicalRF}, have demonstrated efficacy in addressing sketch recognition challenges. More recently researchers have used deep learning for feature extraction which has also improved results \cite{ Peng2019DLFaceDL, Galea2017ForensicFP, Kazemi2018AttributeCenteredLF, Bae2022ExploitingAI, Nagpal2017FaceSM}, and spawned hybrid approaches like \cite{Xu2020MatchingAC}, and \cite{chugh2017transfer}. \\
\subsection{Generative Approaches}
The process of converting images from one modality to another, such as from sketches to photographs, through generative methods was a common strategy employed in photo-sketch recognition tasks. By transforming sketches into photos before matching, applying a straightforward homogeneous face recognition method becomes feasible. Tang and Wang \cite{1238414, 4624272} notably pioneered exemplar-based face sketch synthesis, leveraging a linear transformation assumption between face photos and sketches, using Markov random field models and later, local linear embedding (LLE) \cite{Liu2005ANA}. In \cite{Ouyang2016ForgetMeNotMF}, the authors exploit Gaussian Process regression to deal with both the memory gap and modality gap. However, the efficacy of these approaches in face recognition hinges significantly on the quality of their synthesis results.\\
Following the development of Generative Adversarial Networks (GANs) \cite{Goodfellow2014GenerativeAN}  and cGANs \cite{cGAN}, Isola {\it et al.} proposed a novel method for general image to image translation using cGANs \cite{pix2pix2017,cyclegan}, which gave rise to many sketch generation methods that use GAN's. Recent methods like \cite{HIDA} use Dynamic Normalization (DySPADE) in the generator architecture along with depth maps to supervise the generation. Another recent method called Semi-Cycle-GAN (SCG) \cite{chen2023face-sketch-scg} proposed a semi-supervised approach with a noise-injection strategy to overcome the challenge of missing features in CycleGAN outputs.\\
In AP-DrawingGAN, Yi {\it et al.} \cite{APDrawingGAN2019} used dedicated GANs to generate difficult-to-sketch features like eyes, nose and lips. FSGAN \cite{Fan2021FS2K} extends their approach and introduces a new dataset called FS2K, which has three styles and paired sketch examples. We use this dataset for training and comparison with other methods. In modern approaches like \cite{chan2022drawings}, the authors have used CLIP \cite{clip} along with a geometry-preserving loss to achieve a sketch style that respects the scene's geometry.
\section{Methodology}
In this section, we describe our approach for generating a sketch dataset using diffusion models. Our methodology leverages adapters like ControlNet and IP-Adapter for latent diffusion models to create high-quality, diverse sketch images from mugshot photographs. The core of our approach, which we call CLIP4Sketch, combines the strengths of CLIP and AdaFace embeddings to ensure that generated sketches maintain the identity of the input image while allowing for stylistic variations guided by textual prompts. We begin by outlining the fundamental principles of diffusion models, then detail our specific implementation for sketch generation, including how we incorporate identity preservation and style control into the generation process.\\
\subsection{An Outline of Diffusion Models}
A T2I diffusion model consists of two key elements: the diffusion backbone ($\epsilon_{\theta}$) and the text encoder ($c_{\theta}$). The diffusion backbone undertakes a systematic noise reduction process. Diffusion models operate by sampling noise from a distribution and then iteratively denoising the sample until reaching a final denoised sample, denoted as \( x_0 \). Each step in this process corresponds to a certain noise level, with \( x_t \) representing a blend of signal and noise, where the signal-to-noise ratio is determined by the timestep \( t \). The textual input ($y$) undergoes tokenization and index-based lookup, connecting it to text embeddings ($F_{txt}$) represented as a sequence of vectors. These embeddings are then enhanced by the text encoder, which contextualizes the information, producing refined text embeddings ($E_{txt}$) that capture the essence, intention, and nuances of the textual description. Typically, T2I diffusion models utilize cross-attention layers to leverage the semantic information within $E_{txt}$. The objective function of T2I diffusion models is formulated as:
$$\mathcal{L}=\mathbb{E}_{x, y, \epsilon, t}\left[|| \epsilon_\theta\left(x_t, t, c_\theta\left(F_{t x t}\right)\right)-\epsilon||_2^2\right] $$
\subsection{Face-ID Preservation}
Various strategies have emerged to preserve and personalize face identity of generated images of diffusion models. While some techniques, such as Textual Inversion \cite{Gal2022TI} and DreamBooth \cite{Ruiz2022DreamBoothFT}, necessitate fine-tuning for each new concept, others like IP-Adapter \cite{Ye2023IPAdapterTC} and PhotoMaker \cite{li2023photomaker} offer the advantage of producing identity-consistent generations for multiple subjects without the need for inference time fine-tuning.\\
IP-Adapter and PhotoMaker distinguish themselves by embedding the identity of input reference images into the diffusion process through cross-attention layers. This mechanism directs the diffusion model to generate images consistent with the identities portrayed in the reference images, by using the reference image embedding as condition. However, during empirical evaluation, we found that IP-Adapter lacked the nuanced spatial control necessary to preserve facial identity adequately. For this reason, we decided to use a ControlNet \cite{zhang2023adding} architecture along with an image prompt to achieve more fine-grained spatial control. ControlNet is a neural network structure designed to introduce conditional control to large pretrained text-to-image diffusion models. It leverages source models like LDM and reuses their deep encoding layers \cite{Rombach2021HighResolutionIS,zhang2023adding}, with the intention to construct a robust encoder capable of learning specific conditions like canny edge maps, depth maps and so on. We use a canny edge detector to generate an edge map of the mugshot which we pass in the diffusion pipeline using the ControlNet adapter.
\subsection{CLIP4Sketch Pipeline}
The CLIP4Sketch pipeline begins by taking a mugshot image, denoted as $X_m$, and processing it through a series of steps to generate a corresponding sketch while preserving the identity and introducing various styles. We take inspiration from the decoupled cross attention layer design of IP-Adapter \cite{Ye2023IPAdapterTC} to synthesize sketches with fine-grained control over both the identity and stylistic attributes. \\
We start by encoding the input mugshot image using a pre-trained CLIP model. This model is referred to as $E_{\text{clip}}$, which provides an embedding that captures high level semantic information. Next, in order to ensure that the identity of the individual in the mugshot is preserved in the generated sketches, the pipeline incorporates AdaFace, a robust face recognition model. We denote this model as $E_{\text{fr}}$. The embedding obtained from AdaFace, $E_{\text{fr}}(X_m)$, refines the generated sketch to align closely with the identity characteristics of the input mugshot. The pipeline then concatenates the embeddings from CLIP and AdaFace, resulting in a combined embedding vector, denoted as $C$. We observed empirically that using any of the above-mentioned embeddings by itself does not yield good results; by combining the embeddings from both CLIP and AdaFace, we ensure that the sketches maintain critical identity features along with the style mentioned in textual prompts. This can be represented as:
$$C = [E_{\text{clip}}(X_m), E_{\text{fr}}(X_m)]$$
These combined embeddings are transformed to the same latent space as the textual embeddings using a Projection Network ($P$). 
For style variations, the pipeline utilizes textual captions to guide the stylistic output of the generator. A few such example captions are : ``a software generated sketch of a person", ``a hand drawn sketch of a person". These captions are processed to generate textual embeddings using the CLIP model's text encoder, denoted as $E_{\text{text}}$. The textual embeddings, $E_{\text{text}}(\text{prompts})$, are appended to the combined embeddings ($P(C)$) and passed through the cross-attention layers of our generator. Lets say $\phi$ are weights of image cross-attention layer and $\psi$ are weights of text cross-attention layer, while $\theta$ is the combined weights of the T2I diffusion model. We keep $\psi$ and $\theta$ frozen while only training $\phi$ and our projection network $P$.The diffusion model, which is responsible for generating the sketches, utilizes these final embeddings as condition. We also give a canny edge map to the model as the ControlNet input.  \\
 By inputting different textual descriptions, the cross-attention layers within the diffusion model adjust the sketch generation process to reflect the desired style. This flexibility enables the creation of diverse sketches from the same mugshot, each with unique stylistic elements while preserving the underlying identity. Through this process, CLIP4Sketch effectively bridges the modality gap between photographs and sketches, allowing us to generate a large sketch dataset. So, we generated a synthetic dataset using the proposed CLIP4Sketch, which includes 245,376 sketches of 27,264 identities in four different styles. This synthetic data was crucial in enhancing the our face recognition model's ability to generalize across different sketch styles. Our train-test split strategy ensured that the training set was comprehensive and diverse while the test set accurately reflected real-world scenarios, thereby improving the robustness of our results. In the next few sections, we analyse the potential of our dataset in enhancing the performance of face recognition systems in forensic sketch-to-mugshot matching tasks.
 
\begin{figure*}[t]
% \begin{center}
   \includegraphics[width=1\linewidth]{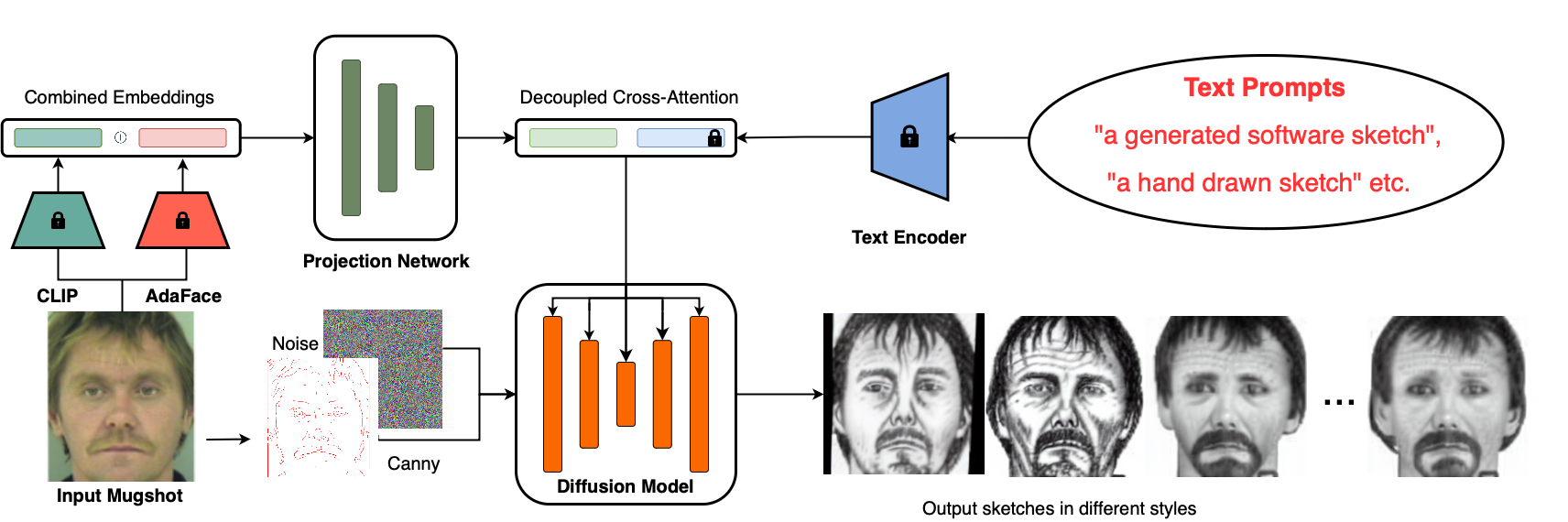}

% \end{center}
   \caption{The CLIP4Sketch pipeline for generating diverse sketches from a mugshot image. A latent diffusion model is employed which combines embeddings from CLIP and AdaFace to preserve identity and uses text prompts to control stylistic variations. We use a canny edge image as controlnet condition along with decoupled cross-attention layers for identity and style conditioning. The output images shown have the following similarity scores 0.2582, 0.2773, 0.2518, 0.2558. }
\label{fig:model}

\end{figure*}
\begin{table*}[htbp]

    \begin{tabular}{|l|p{6.2cm}|c|c|c|}
        \hline
        \textbf{Dataset} & \textbf{Description} & \textbf{Train/Test} & \textbf{No. of Images} & \textbf{No. of Identities} \\
        \hline
        CUHK \cite{4624272}& Hand drawn viewed sketches. & Train & 3,600 & 1,800 \\
        PRIP-Composites \cite{klum} & Viewed and non-viewed sketch composites with mugshot mates. Contains software generated and hand drawn sketches.  & Train & 1,112 & 556 \\
        FS2K \cite{Fan2021FS2K}& Artistic sketch pairs in 3 styles.  & Train &  4,208 &2,104  \\
        TUFTS \cite{tufts}  & Face database for computerized sketches.  & Test &  224 &112 \\
        IIIT-D \cite{Bhatt2012MemeticallyOM} & Forensic and semi forensic pairs of sketches. & Test & 144 & 72 \\
        
        WildSketch \cite{nie2022unconstrained} & Artistic photo sketch pairs  & Test & 1,492 &  796   \\
        Our Synthetic Dataset & Mugshot and sketches in 4 styles.  & Train & 245,376 & 27,264 \\
        \hline
    \end{tabular}\\
    \caption{Summary of the datasets used for training and evaluating the CLIP4Sketch model. The datasets include publicly available sketch and mugshot/photo collections as well as a large synthetic dataset generated using our proposed model.}
    \label{tab:datasets}
\end{table*}

\begin{figure}[t]
% \begin{center}
   \includegraphics[width=0.95\linewidth]{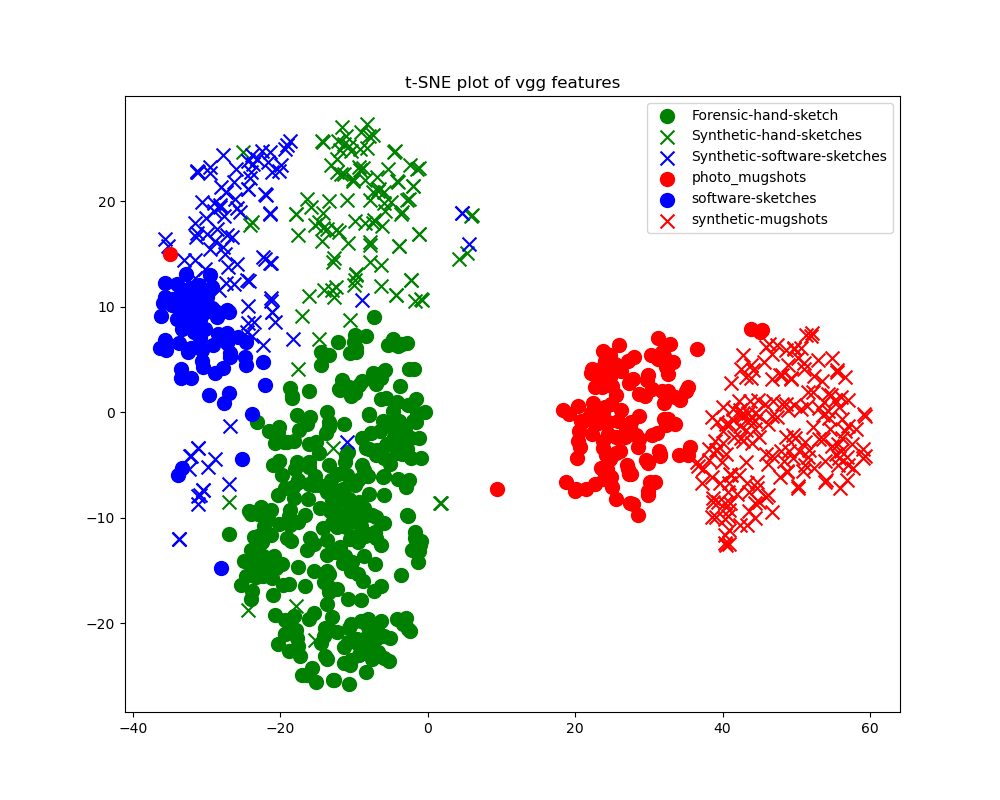}
   \includegraphics[width=0.9\linewidth]{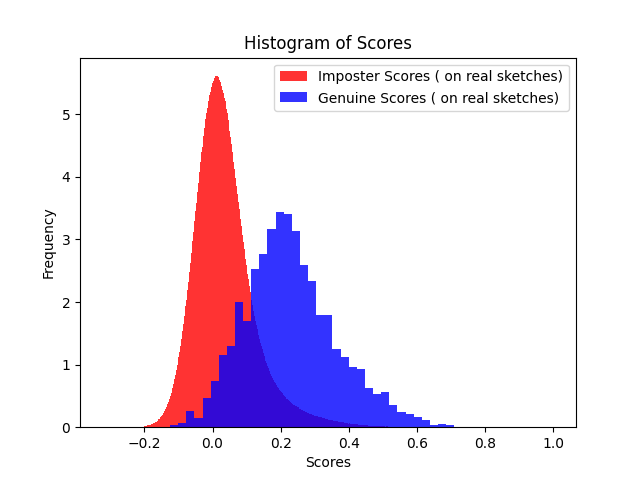}
   \includegraphics[width=0.9\linewidth]{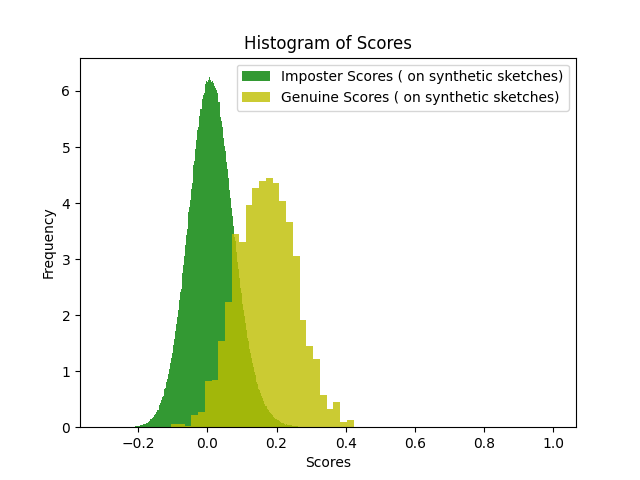}
% \end{center}
   \caption{First, T-SNE plot shows modality gap in sketch images from mugshots and the similarity of our generated sketches and real sketches. The first histogram shows genuine and imposter score distributions for face-sketch pairs in real sketch datasets like CUHK \cite{4624272}, PRIP-Composites \cite{klum} and the second histogram shows the distribution for the dataset generated using our proposed CLIP4Sketch.  } 
\label{fig:vgg}

\end{figure}
\section{Experiments}
To evaluate the effectiveness of our CLIP4Sketch approach and the quality of our synthetically generated sketch dataset, we conducted a series of comprehensive experiments. These experiments were designed to assess multiple aspects of our method, including the realism of the generated sketches, the impact of increasing synthetic data on model performance, and comparisons with existing GAN-based approaches. We also investigated the model's performance in both open-set and closed-set scenarios to provide a thorough assessment of its capabilities. Throughout our experiments, we used various datasets for training and testing, including our synthetic dataset and several publicly available sketch and mugshot datasets. The following subsections detail our experimental setup, methodologies, and findings, providing insights into the strengths and limitations of our approach in enhancing sketch-to-mugshot matching performance.\\
\subsection{Implementation Details}
We conducted our experiments by finetuning pretrained AdaFace models using different datasets. We preprocess the dataset by cropping and aligning faces, resulting in 112 × 112 images. The three augmentations proposed in AdaFace \cite{Kim2022AdaFaceQA} were also applied with a probability of 0.2. We use Adam optimizer \cite{Adam} and a decaying learning rate strategy. It takes us 10 epochs to finetune the model with a batch size of 64. The training was done on 8 A6000 GPU's. \\
\textbf{Datasets :} We use multiple sketch and mugshot datasets to train our sketch generation model, a summary of the datasets used can be found in the table \ref{tab:datasets}. The publicly available datasets included the CUHK Face Sketch Database (CUHK) \cite{4624272}, PRIP-Composites \cite{klum}, FS2K \cite{Fan2021FS2K}, TUFTS Face Database \cite{tufts}, IIIT-D Forensic Sketch Database \cite{Bhatt2012MemeticallyOM}, and WildSketch \cite{nie2022unconstrained}. The CUHK, PRIP-Composites, and FS2K datasets, comprising a total of 8,912 sketches representing 4,460 unique identities, were used exclusively for training the CLIP4Sketch pipeline. For testing the FR system, we employed the TUFTS Face Database, IIIT-D Forensic Sketch Database, and WildSketch, collectively offering 1860 sketches of 980 identities, to provide a realistic evaluation of our model's performance.\\
\begin{table*}

    \begin{tabular}{|p{7.8cm}|c|c|c|c|}
        \hline
        \textbf{Model \& Dataset} & 
        \multicolumn{4}{|c|}{\textbf{ TAR (\%) @ FAR=0.1\% }}\\
        \hline
        
          &CASIA&
         WildSketch & 
         IIIT-D Sketches &
         TUFTS\\
        \hline
        Pretrained Adaface  & 94.10 & 66.10 &55.80 & 14.41\\
        Adaface finetuned on 25\% of synthetic data  & 53.20 $\pm 1.5$  & 82.70 $\pm{2.28}$  & 44.80 $\pm 5.60$ & 18.44 $\pm 3.50$\\
        Adaface finetuned on 50\% of synthetic data & 46.00 $\pm 3.3$  & 84.03 $\pm 1.25$ & 49.61 $\pm 7.70$  & 19.11 $\pm 4.30$\\
        Adaface finetuned on 75\% of synthetic data &  39.40 $\pm 1.5$   & 86.67 $\pm 1.83$ & 60.5 $\pm 5.40$ & 19.40 $\pm 3.68$\\
        Adaface finetuned on 100\% of synthetic data &   35.10  & 85.88 & 58.80 & 20.75\\
        \hline
    \end{tabular}
        \\
    \caption{Performance comparison of AdaFace model finetuned on varying amounts of synthetic data. The table shows True Accept Rate (TAR) at a False Accept Rate (FAR) of 0.1\% across different datasets. The upward trend in sketch datasets shows the potential of our approach while high variance in datasets like TUFTS and IIIT-D shows the inherent challenges in them. }
    \label{tab:exp2}
\end{table*}

\begin{table*}[htbp]
    \begin{tabular}{|p{7.7cm}|c|c|c|c|}
        \hline
        \textbf{Model \& Dataset} & \textbf{Closed-set Rank-1 (\%)} & 
         \textbf{Open-set FNIR (\%) @FPIR=2\%}\\ 
        \hline
        Pretrained Adaface  & 72 & 51 \\
        Adaface finetuned on Face-Sketch-SCG \cite{chen2023face-sketch-scg}  & 14   & 97  \\        Adaface finetuned on InformativeDrawings \cite{chan2022drawings} &  59   & 88  \\
        
        Adaface finetuned on our dataset & \textbf{85}  & \textbf{33}  \\

        \hline
    \end{tabular}\\
    
    \caption{Comparison of closed-set Rank-1 accuracy and open-set FNIR@FPIR=2\% between our synthetically generated dataset and another dataset made using InformativeDrawings  \cite{chan2022drawings} which use GANs for sketch generation. The gallery size was 10,000, and the probe set had 980 identities.}
    \label{tab:exp3}
\end{table*}

\subsection{Realism of Generated Sketches}
To validate the realism and quality of our CLIP4Sketch generated sketches, we conducted a comprehensive comparative analysis between the generated sketches and real forensic sketch images. This analysis focused on two key aspects: the distribution of similarity scores and feature space representation. Using a pre-trained AdaFace \cite{Kim2022AdaFaceQA} network, we extracted features from both sets of sketches and calculated scores for genuine matches (sketches matched with their corresponding mugshots) and imposter matches (sketches matched with non-corresponding mugshots). Our analysis revealed that the score distributions of the generated sketches closely resemble those of the real forensic sketches, indicating a high level of realism in the synthetic images.\\
Additionally, we use t-SNE \cite{tSNE} to visualize the similarity between generated and real sketches in the VGG latent space. Our t-SNE plots demonstrated that the clusters of generated sketches were highly similar to those of real sketches. This clustering behaviour indicates that our CLIP4Sketch-generated sketches share similar feature characteristics with real forensic sketches, further reinforcing our quantitative findings from the similarity score analysis. In figure \ref{fig:vgg} the t-SNE visualization effectively validated the realism of our DDPM-generated sketches, supporting their potential use in improving sketch-to-mugshot matching performance.\\
\subsection{Effect of Size of Synthetic Dataset}
To investigate the impact of increasing synthetic data on sketch-to-face matching performance, we conducted an experiment varying the proportion of synthetic data in our training dataset. We split the dataset into four compositions: 25\%, 50\%, 75\%, and 100\% synthetic data, corresponding to 6,816, 13,632, 20,448, and 27,264 unique identities, respectively. We evaluated the impact of increasing synthetic data on both sketch-to-face matching and face-to-face matching performance. For face-to-face matching, we used the CASIAWebFace test set \cite{Webface} to assess how the increase in synthetic data affects the model's performance on traditional face recognition tasks.\\
Our quantitative analysis revealed interesting trends. As the proportion of synthetic data increased, we observed a general improvement in sketch-to-face matching performance across multiple test datasets (WildSketch, IIIT-D Sketches, and TUFTS). This suggests that the synthetic data effectively helps the model learn features that bridge the gap between sketch and photo domains.
However, we also noted a decline in face-to-face matching scores on the CASIA dataset as the proportion of synthetic data increased. This decrease highlights a trade-off: while finetuning with more synthetic data improves sketch-to-face matching, it may lead to a degradation in the model's ability to recognize real faces in traditional face recognition tasks.
These findings show the complex relationship between quantity of cross modality data while finetuning and general model performance. While synthetic sketch data is crucial for improving performance on specialized tasks like sketch-to-face matching, increasing its proportion may impact general face recognition capabilities. This suggests that in practical applications, careful consideration should be given to the balance of synthetic and real data in the training set, depending on the specific requirements of the task at hand.

\begin{figure}[t]
% \begin{center}
   \includegraphics[width=1\linewidth]{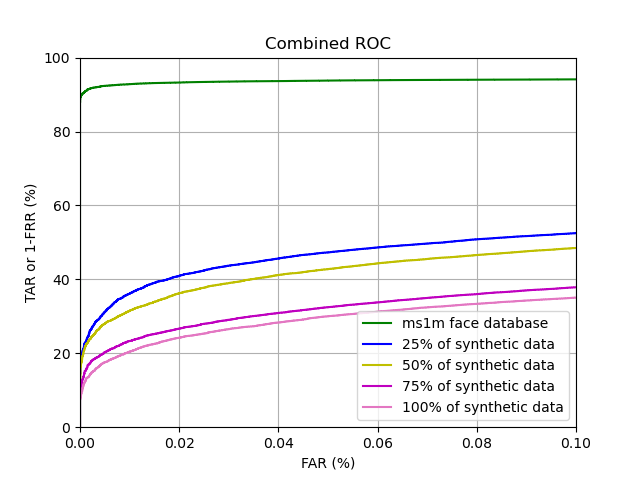}
   \includegraphics[width=1\linewidth]{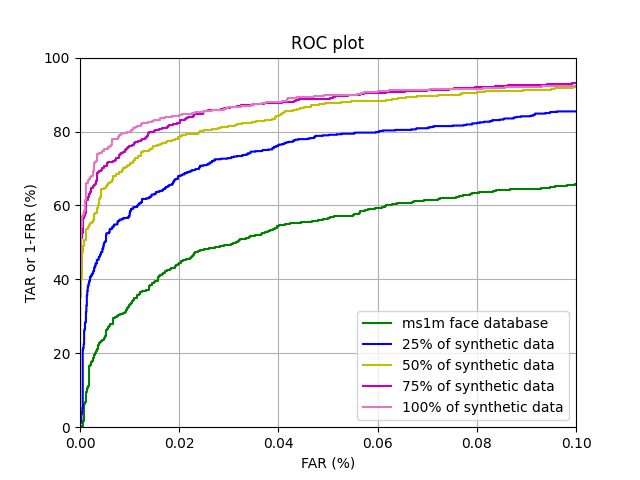}
% \end{center}
   \caption{ROC plots that show how performance drops for face to face matching because of using more sketch data, while an upwards trend is visible for face to sketch matching as we use more data.  } 
\label{fig:exp2}

\end{figure}
\begin{figure}[t]
% \begin{center}
   \includegraphics[width=1\linewidth]{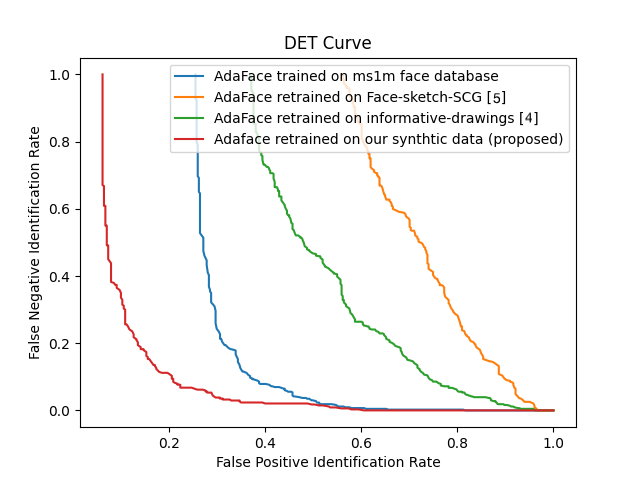}

% \end{center}
   \caption{DET plot to show open set performance of Adaface when trained on our dataset, compared against other GAN generated datasets. This shows the potential of diffusion models in generating synthetic datasets.}
\label{fig:tasets}

\end{figure}

\subsection{Comparison with GAN-Generated Datasets}

In this experiment, we compare the performance of the face recognition model trained on our dataset generated using the proposed CLIP4Sketch with models trained on datasets generated by InformativeDrawings \cite{chan2022drawings} and Face-Sketch SCG \cite{chen2023face-sketch-scg}. We use their Anime-Sketch style for our evaluations. Specifically, we evaluate the models on both open-set and closed-set scenarios to assess their performance comprehensively. Our gallery consists of 10,000 mugshots, and the probes consist of the test datasets mentioned in the datasets table \ref{tab:datasets}. As shown in Table \ref{tab:exp3}, our model trained on the DDPM-generated dataset outperformed the model trained on the Informative Drawings dataset in both open-set and closed-set scenarios. Specifically, our model achieved an open-set FNIR of 33\% and a closed-set rank1 accuracy of 85\%, compared to 88\% and 59\%, respectively, for Informative Drawings \cite{chan2022drawings}, 97\% and 14\% for Face-SCG \cite{chen2023face-sketch-scg}. Though there is much scope for improvement, these results show the potential of synthetic datasets in cross-modal face recognition.  
\subsection{Failure Cases}
\begin{figure}[t]
% \begin{center}
   \includegraphics[width=0.9\linewidth]{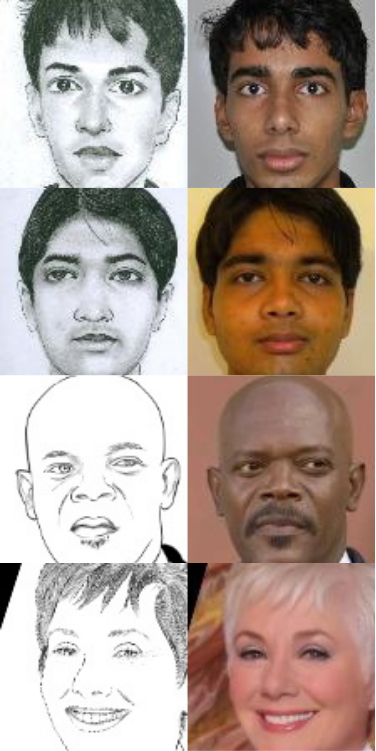}

% \end{center}
   \caption{A few examples of sketches that were incorrectly matched by our trained model. First 2 rows show examples from IIIT-D \cite{Bhatt2012MemeticallyOM} dataset, and last 2 rows show examples from the WildSketch \cite{nie2022unconstrained} dataset. }
\label{fig:failure}

\end{figure}

Despite the significant improvements achieved by CLIP4Sketch in enhancing sketch-to-mugshot matching performance, there remain certain cases as shown in figure \ref{fig:failure}, where our face recognition model, trained on the synthetic sketch dataset, struggles to accurately match faces with forensic sketches. These failure cases highlight the inherent challenges posed by the substantial modality gap between photographs and forensic sketches, especially in IIIT-D \cite{Bhatt2012MemeticallyOM} and TUFTS \cite{tufts} datasets. This suggests that simply increasing the dataset size alone may not fully bridge this divide.
Forensic sketches, particularly those created by witnesses or forensic artists based on recollections and descriptions, can deviate significantly from the subject's actual facial features and proportions. These deviations introduce complexities that our model may only sometimes be able to overcome, even with the aid of the synthetic data.
In contrast, our model performs better when matching artistic sketches, which often adhere more closely to the actual facial structures and proportions. However, the real challenge lies in the forensic domain, where sketches can be highly subjective and prone to distortions.

\section{Conclusion}

In this work, we introduced CLIP4Sketch, a novel approach leveraging Denoising Diffusion Probabilistic Models (DDPMs) tailored to generate diverse sketches from mugshot images. By combining the strengths of CLIP and AdaFace embeddings, our method ensures that the generated sketches retain the identity of the input image while allowing for diverse stylistic variations guided by textual captions.
Our extensive experimental evaluations demonstrate that incorporating synthetic sketches generated by CLIP4Sketch into sketch training datasets significantly enhances the performance of face recognition models in sketch-to-mugshot matching tasks. The utilization of synthetic data in this context not only addresses the quantitative needs of modern FR evaluation but also alleviates privacy concerns associated with large-scale collection and use of authentic facial images. Moving forward, this approach can be extended to other domains requiring cross-modal image generation and matching, paving the way for more versatile and practical solutions in facial recognition technology.

% \begin{figure*}
% \begin{center}
% \fbox{\rule{0pt}{2in} \rule{.9\linewidth}{0pt}}
% \end{center}
%    \caption{Example of a short caption, which should be centered.}
% \label{fig:short}
% \end{figure*}

%------------------------------------------------------------------------

%-------------------------------------------------------------------------

{\small
\bibliographystyle{ieee}
\bibliography{egbib}
}

\end{document}